\documentclass[conference]{IEEEtran}
\IEEEoverridecommandlockouts
\usepackage{cite}
\usepackage{amsmath,amssymb,amsfonts}
\usepackage{algorithmic}
\usepackage{graphicx, subfigure}
\usepackage{textcomp}
\usepackage{xcolor}
\usepackage{url}
\def\BibTeX{{\rm B\kern-.05em{\sc i\kern-.025em b}\kern-.08em
    T\kern-.1667em\lower.7ex\hbox{E}\kern-.125emX}}

\newcommand{\myparagraph}[1]{\vspace{0.02cm} \noindent \textbf{#1}}

\begin{document}

\title{PyTAG: Challenges and Opportunities for Reinforcement Learning in Tabletop Games}

\author{\IEEEauthorblockN{Martin Balla, George E.M. Long, Dominik Jeurissen, James Goodman, Raluca D. Gaina, Diego Perez-Liebana}
\IEEEauthorblockA{\textit{Queen Mary University of London} \\
\{m.balla, g.e.m.long, d.jeurissen, james.goodman, r.d.gaina, diego.perez\}@qmul.ac.uk}
}


\IEEEoverridecommandlockouts

\IEEEpubid{\makebox[\columnwidth]{979-8-3503-2277-4/23/\$31.00~\copyright2023 IEEE \hfill} \hspace{\columnsep}\makebox[\columnwidth]{ }}

\maketitle

\IEEEpubidadjcol

\begin{abstract}
In recent years, Game AI research has made important breakthroughs using Reinforcement Learning (RL). Despite this, RL for modern tabletop games has gained little to no attention, even when they offer a range of unique challenges compared to video games. To bridge this gap, we introduce PyTAG, a Python API for interacting with the Tabletop Games framework (TAG). TAG contains a growing set of more than 20 modern tabletop games, with a common API for AI agents. We present techniques for training RL agents in these games and introduce baseline results after training Proximal Policy Optimisation algorithms on a subset of games. Finally, we discuss the unique challenges complex modern tabletop games provide, now open to RL research through PyTAG.

\end{abstract}


\section{Introduction}
In recent years, Reinforcement Learning (RL) has made various breakthroughs for playing board games like Go~\cite{schrittwieser2020muzero} or Chess~\cite{alphazero}, as well as video games like ``Starcraft''~\cite{alphastar}. By repeatedly playing games millions of times, these algorithms are able to learn powerful complex strategies and solutions to various problems, in single-player or adversarial settings, which rival or even exceed the abilities of champion human players. However, most RL applications focus on video games, classical board or card games, as opposed to modern Tabletop Games (TTGs). This is due to the fact that, although there are some implementations of TTGs, no framework provides a unified API for playing a wide range of modern TTGs using RL, facilitating wide adoption of research on these games.

In contrast to classical board games, modern TTGs like ``Settlers of Catan'' or ``Pandemic'' are typically played with more than $2$ players, which raises the problem of multi-agent dynamics: how can one's strategy be optimised according to the behaviour of multiple other intelligent entities acting upon the same environment? In most TTGs, players are competing against each other; however, others are collaborative (e.g. ``Pandemic''), where players have to work together and coordinate their actions to defeat either the game or another group of players. Modern TTGs also include aspects of high uncertainty, hidden information, and very large, diverse and complex state and action spaces. Modern TTGs present a wide variety of challenges and a benchmark for RL to be explored. 

The Tabletop Games framework (TAG) implements a large set of modern TTGs with a common interface for AI agents. TAG is implemented in Java, which allows running games with fast Forward Models; this is essential for Statistical Forward Planning (SFP) agents, which run in real-time and use the Forward Models to generate simulations of possible futures, in order to build statistics describing best strategies. SFP methods are fast to run, without any training required, and are able to produce good performance in a wide variety of games, with limited or zero expert knowledge. However, these methods heavily rely on the existence of a Forward Model (which may not be available for many real-world problems), and can struggle to reach optimum play due to the constraints of fast decision-making in real-time. 

The integration of RL agents into TAG balances out these drawbacks, allowing for more complex strategic behaviours to be learned, amplifying the usefulness of TAG as a framework for not only developing high-performing AI players, but also for the study and improvement of the design of modern TTGs. So far, running RL agents in TAG has been challenging due to the lack of libraries required for implementing efficient RL agents in Java. Further, as TAG development was focused on SFP agents playing games, TAG is highly tailored to the needs of these algorithms. Therefore, several modifications to the framework are needed to allow the addition of RL agents. 

This work introduces PyTAG, a Python interface for interaction with TAG~\cite{gaina2020tag}. The communication between TAG and PyTAG is done by sharing the memory locations between Java and Python, which keeps the games running fast and efficiently. Further, this paper explores the challenges that arise when applying RL agents to modern TTGs through PyTAG, and proposes solutions and guidance for future research. The focus of this paper is not to outperform the existing SFP agents in TAG, but rather to take steps towards tackling the challenges that modern TTGs raise for training RL agents, in order to further diversify the library of AI players available in TAG for the in-depth analysis of modern TTGs. All our code (baseline agents, wrappers, python interface) is available on Github\footnote{\url{https://github.com/martinballa/PyTAG}}. 


\section{Background}
\subsection{Tabletop Games framework}
The Tabletop Games framework (TAG)~\cite{gaina2020tag} is designed for research on modern TTGs. It provides a common API for SFP agents, various tools for automatic game optimisation, and metrics for evaluating both agents and games. To this extent, TAG provides a set of components, templates and interfaces to quickly implement new games, and a standard API for creating AI agents to play them. At the time of writing, TAG includes a collection of over 20 modern TTGs, including ``Settlers of Catan'', ``Pandemic''~\cite{gaina2022tag}, and ``Terraforming Mars''~\cite{gaina2021tag}. All agents have access to the Forward Model of the game, which can be used to simulate potential future states in the game, when provided with a previous game state and action. Several general-purpose SFP agents are implemented in TAG, such as Monte Carlo Tree Search (MCTS~\cite{browne2012survey}) and Rolling Horizon Evolutionary Algorithm (RHEA~\cite{gaina2021rolling}). 


\subsection{Reinforcement Learning}
Reinforcement Learning (RL) is an area of machine learning which focuses on learning a policy to solve sequential decision-making problems. The combination of RL with Deep Neural Networks has achieved superhuman performance on a range of domains~\cite{schrittwieser2020muzero}\cite{alphastar}, but most works focused on either video games (Atari, ``Starcraft'') or classical board games (Chess, Go). Modern board games share many similarities with video games, but they are relatively unexplored with RL due to the lack of research-friendly software implementations.


Formally, RL is defined as a Markov Decision Process where at each time step $t$, the agent receives a state $s_t$ from the state space $\mathcal{S}$. In each state, the agent executes an action $a_t$ from the action space $\mathcal{A}(s_t)$ following its policy $\pi(a_t|s_t)$. After executing $a_t$, the state transitions to the next state $s_{t+1}$, according to the dynamics of the environment $\mathcal{P}(s_{t+1}|s_t,a_t)$. The agent receives a scalar reward $r_t$ from the function $\mathcal{R}(s_t,a_t)$, representing the objective of the environment. The agent aims to maximise the return (cumulative rewards) $R_t = \sum_{k=0}^{\infty}\gamma{}^kr_{t+k}$, where $\gamma{}$ is the discount factor weighting long-term future rewards.

Model-free RL algorithms are typically classified according to whether they represent a policy based on a value function (Q-values), or directly (policy gradients). Policy gradient methods are better suited to deal with large action spaces that TTGs provide. Actor-critic algorithms are an extension to policy gradient algorithms, as they represent a policy $\pi_\theta{}(s_t|a_t)$  and a value function $Q_\omega{}(s_t,a_t)$ separately. The policy's parameters $\theta$ are trained to maximise the return, while the critic is trained to predict the expected return under the learned policy. Actor-critic methods use the critic to stabilise learning. Proximal Policy Optimisation (PPO)~\cite{schulman2017proximal} is one of the most popular actor-critic algorithms, due to its simplicity and high performance across various benchmarks. 


\section{Related Work}

\subsection{AI and Board Games}

Games are one of the most prolific benchmarks for testing progress in AI research. In particular, board and card games have been used often as testbeds for new algorithms. A large amount of research on Game AI has focused on video-games~\cite{bellemare2013arcade} and classic board games, such as chess~\cite{alphazero}, go~\cite{schrittwieser2020muzero} and poker~\cite{moravvcik2017deepstack}. TTGs are considerably richer domains, compared to classic board games, where players need to manage hidden information, boards, cards, resources and stochasticity. Games can be competitive, cooperative or usually a mixture of these, with coalitions of players forming and changing over the course of a game~\cite{engelstein2022building}. Creating AI players for TTGs is challenging: AI agents must have the ability to cooperate or compete, making strategic and tactical decisions in unclear situations where crucial information is missing.

Early work on TTGs used MCTS agents to deal with multi-player and non-deterministic games such as ``Settlers of Catan''\cite{szita2009monte} or ``Magic: The Gathering''\cite{ward2009monte}. MCTS was also used in ``Risk''\cite{gibson2010automated} to aid the drafting of game territories by players at the start of the game, or in ``The Resistance''\cite{cowling2015emergent} to create agents able to bluff effectively. Some research has focused on identifying strategies to play better games such as ``Ticket to Ride''\cite{de2017ai} or ``One Night Ultimate Werewolf''\cite{eger2019study}. 

Research in the card game ``Hanabi'', which focuses on creating collaborative agents, has been coined as the ``new frontier for AI research''\cite{bard2020hanabi} for Human-AI cooperation. The \textit{Hanabi Learning Environment} implements an RL API for this game, which is particularly interesting for RL research due to its cooperative, multi-agent and imperfect information nature. Players must act as a team and provide hints to each other to avoid making errors. Existing techniques like self-play are problematic for this kind of game, as players need to adapt to their team and cannot assume that each player plays optimally or follows the same strategy.

Other recent and remarkable works have been achieved in the domain of TTGs. For instance, \textit{Player of Games}, a general-purpose algorithm that combines guided search, self-play learning, and game-theoretic reasoning, achieved strong empirical performance in the imperfect information game ``Scotland Yard''~\cite{schmid2021player}. We further highlight the recent work by Meta which achieved human-level play for the TTG ``Diplomacy'', showing the capacity of large language models and self-play algorithms~\cite{meta2022human}.

\subsection{Multi-game AI Frameworks}

However, a large proportion of traditional and current Game AI research, including many of the works mentioned above, use different algorithms and software frameworks. This generally makes comparing results and transferring knowledge between games difficult and time-consuming. For this reason, multiple benchmarks have been proposed in the last decade, aiming to provide a common API for AI agents, so new algorithms can be tested in several games at once.
Popular video-game frameworks that follow this philosophy are the Arcade Learning Environment (ALE;~\cite{bellemare2013arcade}) and the General Video Game AI Framework (GVGAI;~\cite{GVGAIsurvey} for simple two-dimensional arcade games. Recently, Griddly~\cite{bamford2022griddly} and Stratega~\cite{dockhorn2020stratega} provide a common API for creating general AI players for multiple real-time and turn-based strategy games. 

In the domain of TTGs, \textit{RLCard} is a framework for evaluating RL and search algorithms in card games, such as Blackjack or UNO~\cite{zha2019rlcard}. \textit{OpenSpiel}~\cite{lanctot2019openspiel} provides a wide range of games and algorithms for research in RL and Search algorithms in single- and multi-agent games, including a wide range of classical board games like ``Connect 4'' and Tic Tac Toe. Finally, \textit{Ludii}~\cite{piette2019ludii} is a framework and language for defining and playing a wide range of games, including board games. \textit{Ludii} has been applied for general game playing and generating new games, and it also supports an API for RL algorithms. While these are useful and interesting frameworks for AI research, none of them have the capacity to implement complex and rich TTGs.

Three aspects motivate the work described in this paper: the increasing popularity of TTGs as benchmarks for Game AI research, the current successes of (Deep) RL in the field, and the lack of a framework that provides a common interface for creating RL agents that play multiple TTGs through a common interface.

\section{PyTAG}

PyTAG is a python interface for interacting with the TAG framework, with the primary purpose of supporting RL agents on the collection of games TAG provides. As RL agents require a large amount of experience, slow environments are unsuitable for large-scale experiments. TAG has already shown high running speeds~\cite{gaina2020tag}, but Java has little support for libraries that allow efficient implementation of RL agents, which leads to the requirement to implement a python API. So far, implementing new agents was only available in Java. However, with PyTAG, researchers can implement new agents in Python as well, which lowers the entry barrier for using TAG for RL research.
PyTAG was developed following the philosophy of keeping the python interface customisable, including its potential use for non-RL purposes. The communication between the programming languages is done by sharing memory locations of objects between Java and Python. As a result, the communication of observations and actions remains fast, which is required to train RL agents. 

One challenge that comes with tabletop games, compared to other multi-game frameworks, is that all games are unique in terms of observation spaces, action spaces and rewards, making the implementation of a general interface for RL challenging. We have selected a few games that present an assorted set of challenges for this work and extended them with interfaces to PyTAG. 
PyTAG supports OpenAI's gym~\cite{brockman2016openai} interface, which allows using common wrappers to run multiple environments in parallel for training, and to modify the observation spaces, action spaces and reward functions. 

\subsection{Observations}

The first PyTAG interface converts the information extracted from a game state, to a fixed format that can be used as input to a neural network. Two options are possible: from the Java side, we return either a \textbf{vectorised observation} of the state, or a \textbf{JSON object}. The former is the simpler and faster approach, but it is limiting for games with complicated observation spaces such as ``Settlers of Catan'' or ``Terraforming Mars''. Working with JSON objects allows more flexibility and control over what is given to the Python agent for action selection, but the extra conversions may slow down training. 


\subsection{Actions}

The second PyTAG interface deals with how actions are used by the agent. By default, TAG computes legal actions in a game state, which is sufficient for SFP agents to make decisions. Unfortunately, RL agents typically expect a fixed action space, with known dimensions at initialisation. The action spaces in tabletop games are rarely enumerable: they are dynamic and highly dependent on the game state. To generate a fixed action space, we implemented \textbf{action trees} for all games in our experiments, similar to the work by~\cite{bamford2021generalising}. Each leaf node in the action tree represents an action, and the intermediate nodes represent action categories. 

The shape of the action tree is fixed when the game is initialised, but the available actions in the tree are updated at each step. The action tree can be used to determine which actions are legal in the current state, which we refer to as the \textbf{action mask}. The action mask is a boolean vector with the same size as the number of possible actions in the game. This can be used to filter predicted Q-values or policy logits before action selection. For Q-values, the unavailable actions are replaced by a small value (i.e. $1e^{-8}$), so, at action selection, $argmax_a Q(s, a)$ will not pick them. In the case of policy gradients, the masking is done on the predicted policy logits: the unavailable units are replaced by a small negative value, so, when the action probabilities are calculated, those actions produce $p=0.0$ and are not chosen. 

A limitation to our experiments is the lack of direct use of the action tree structures, due to the complexities of using irregularly shaped trees. Deep Learning libraries are designed to speed up computation on regular-sized tensors only. Some of the games do satisfy this requirement: Tic Tac Toe and ``Stratego'' both have regularly shaped action trees, but the other games do not. We hypothesise that action trees could be used to better handle the action spaces with specialised Neural Network architectures, or by using regressive action selection.

\section{Experiments}
This section presents baseline experiments using PPO on a subset of TAG games. Our main aim is to demonstrate the use of PyTAG, and propose solutions to the challenges of complex observation and action spaces.

\begin{figure*}[t!]
    \centering
    \subfigure{\includegraphics[width=0.24\textwidth]{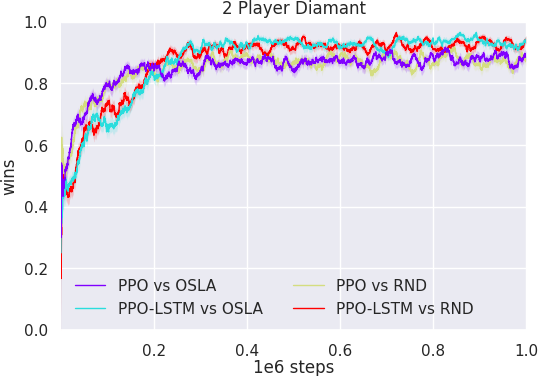}} 
    \subfigure{\includegraphics[width=0.24\textwidth]{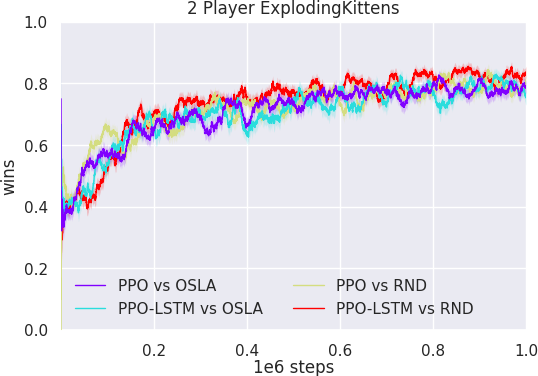}} 
    \subfigure{\includegraphics[width=0.24\textwidth]{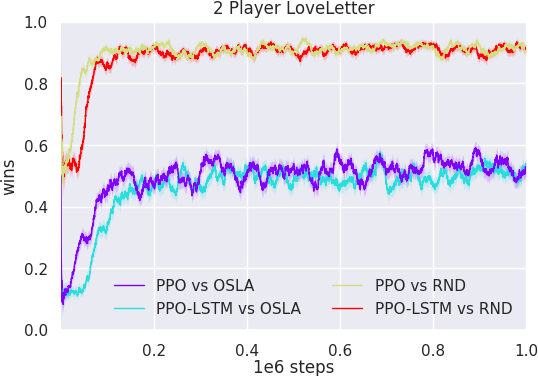}} 
    \subfigure{\includegraphics[width=0.24\textwidth]{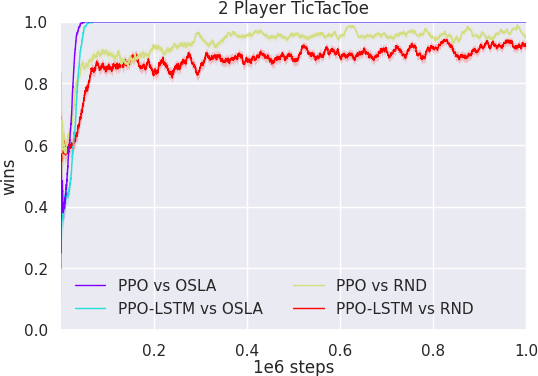}} 

    \caption{2-player results in ``Diamant'', ``ExplodingKittens'', ``LoveLetter'' and Tic Tac Toe (from left to right). Graphs show the running mean of episodic win-rates, averaged across all seeds. Shaded areas show the standard error.}
    \label{fig:2p_win}
\end{figure*}

\begin{figure*}[t!]
    \centering
    \subfigure{\includegraphics[width=0.32\textwidth]{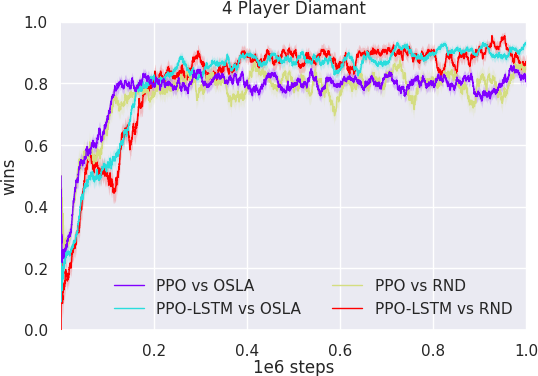}} 
    \subfigure{\includegraphics[width=0.32\textwidth]{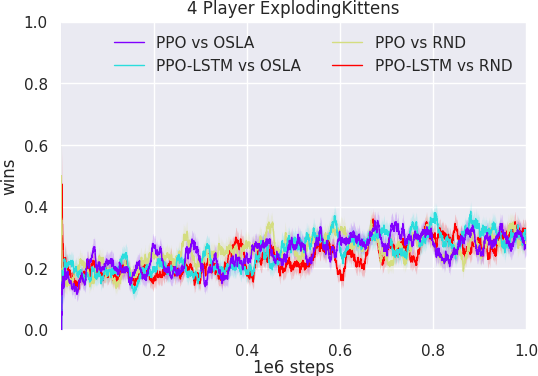}} 
    \subfigure{\includegraphics[width=0.32\textwidth]{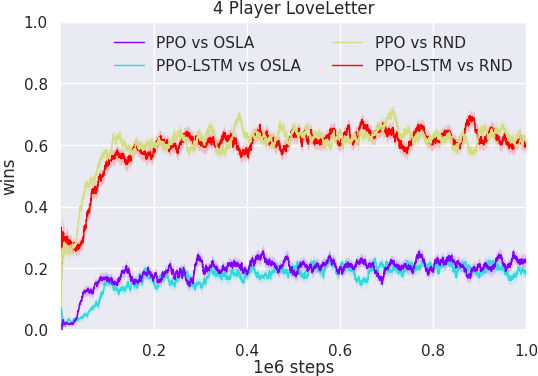}} 
    \caption{4-player results in ``Diamant'', ``ExplodingKittens'' and ``LoveLetter'' (from left to right), showing the running mean win-rates across training. }
    \label{fig:4p_win}
\end{figure*}

\begin{figure*}[t!]
    \centering
    \subfigure{\includegraphics[width=0.32\textwidth]{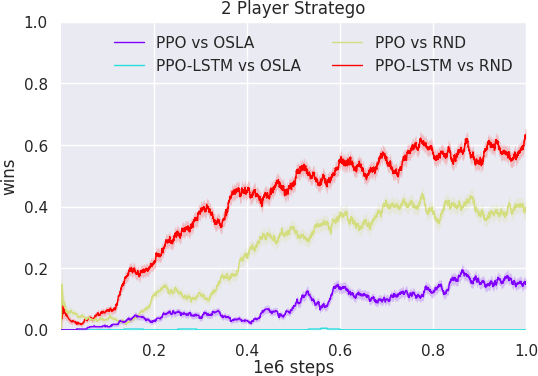}} 
    \subfigure{\includegraphics[width=0.32\textwidth]{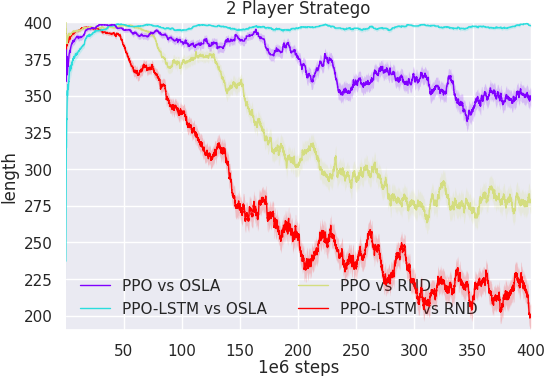}} 
    \subfigure{\includegraphics[width=0.32\textwidth]{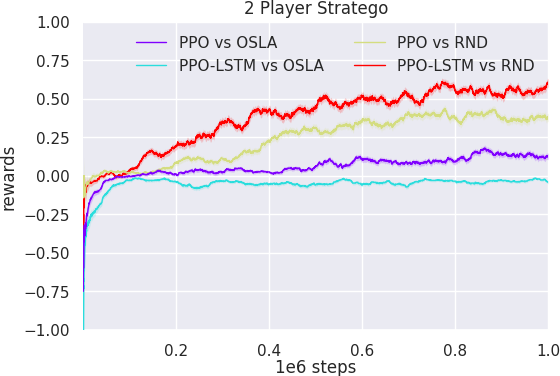}} 

    \caption{``Stratego'' results showing the running mean win-rate (left), episode lengths (middle) and rewards (right).}
    \label{fig:stratego}
\end{figure*}

\subsection{Experimental setup}

TAG includes several AI players that were used as opponents for training the RL agents. We employed \textit{Random} and \textit{One-step look-ahead} (OSLA). Random picks an action from all available with uniform probability, while OSLA tries all available actions, using the Forward Model, to determine the best next action, with a look-ahead of one move. RHEA and MCTS agents were not used for training, due to the considerable overhead required for each decision.

Due to challenging action spaces and action masking, actor-critic RL algorithms are better suited to develop our baseline agent. We have adapted the PPO algorithm~\cite{schulman2017proximal} from CleanRL~\cite{huang2022cleanrl}. As most games have hidden information, we have adapted a version of PPO with a Long Short Term Memory (LSTM;~\cite{hochreiter1997long}) network as the second baseline, to add memory for decision-making. To handle the observations received from TAG, we modified the Neural Network (NN) architecture for compatibility with the observation vectors per game. The NNs have $2$ fully connected layers and $2$ heads to output the action logits and the values for all games with vector observations. For training, we used the default parameters from CleanRL. For ``Stratego'', we used an extra convolutional layer to extract spatial information from the board. 

All agents were trained for $1$ million time steps across $4$ seeds on a machine with $8$ CPU cores and $1$ GPU (2080 TI). For all games, we used $8$ parallel instances of TAG with synchronous steps, employing OpenAI gym's vectorised environment wrapper. As the reward function, we used win/loss signals for all games: $+1$ for winning the game, $0$ for a tie and $-1$ for losing the game. Most of the games are fairly short, with episode lengths of less than $40$ steps on average (just counting our agents' decisions). Only ``Stratego'' has longer episodes, with $600-700$ steps on average, while TAG's implementation declares a draw after $400$ steps. Our experiments were done on $2$- and $4$-player versions of the selected games. When training on $2$-player games, we used $1$ PPO vs $1$ baseline agent. In $4$-player games, we used $1$ PPO vs $3$ instances of the selected baseline agent.
For all games, we measure rewards, win rates, length of episodes, and Frames Per Second (FPS; which measures the rate of the interactions of the agent, including processing observations, optimising the agent, and waiting for the opponents to act). For the number of steps, only the times when PPO acts are counted. 

\begin{table*}[!t]
\begin{center}
\caption{Results summary showing mean and standard error when training PPO (without LSTM). Wins, returns and episode lengths are calculated in the last $100$ episodes, while the Frames Per Second (FPS) is calculated throughout training (including optimising PPO and waiting for the opponents to act). }

\begin{tabular}{|l|l|l|l|l|l|l|l|}
\hline
\textbf{Opponent} &\textbf{ Number of Players} &              \textbf{Game} &   \textbf{Wins} &   \textbf{Returns} &    \textbf{Episode Lengths} &      \textbf{Mean FPS} \\
\hline  
         Random & 2 &         Tic Tac Toe &  0.96 (0.02) &   0.94 (0.03) &     3.19 (0.04) &    754.52 (0.79) \\
         Random &  2 &           Diamant &  0.82 (0.04) &   0.65 (0.08) &    18.22 (0.31) &   1950.20 (9.35) \\
         Random & 2 &        LoveLetter &  0.93 (0.03) &   0.86 (0.05) &    22.72 (0.49) &   1880.61 (2.70) \\
         Random & 2 &  ExplodingKittens &  0.74 (0.04) &   0.48 (0.09) &    32.70 (0.98) &   1777.53 (3.79) \\
         Random & 2 &          Stratego &  0.26 (0.05) &   0.24 (0.05) &  330.91 (13.33) &    184.20 (0.60) \\
\hline         
         Random & 4 &           Diamant &  0.86 (0.03) &   0.72 (0.07) &    22.09 (0.33) &  1431.56 (13.96) \\
         Random & 4 &        LoveLetter &  0.59 (0.05) &   0.18 (0.10) &    20.91 (0.47) &   1455.32 (1.28) \\
         Random & 4 &  ExplodingKittens &  0.35 (0.05) &  -0.30 (0.10) &    21.10 (0.85) &   1731.85 (1.85) \\

\hline
         OSLA & 2 &         Tic Tac Toe &  1.00 (0.00) &   1.00 (0.00) &    3.00 (0.00) &   703.99 (0.52) \\
         OSLA & 2 &           Diamant &  0.85 (0.04) &   0.70 (0.07) &   18.55 (0.28) &  1931.23 (6.56) \\
         OSLA & 2 &        LoveLetter &  0.52 (0.05) &   0.04 (0.10) &   25.14 (0.67) &  1466.61 (1.89) \\
         OSLA & 2 &  ExplodingKittens &  0.77 (0.04) &   0.54 (0.08) &   34.15 (1.03) &  1758.99 (3.36) \\
         OSLA & 2 &          Stratego &  0.02 (0.01) &  -0.01 (0.02) &  390.07 (5.27) &    24.57 (0.11) \\
\hline
         OSLA & 4 &           Diamant &  0.75 (0.04) &   0.51 (0.09) &   21.58 (0.33) &  1212.58 (9.69) \\
         OSLA & 4 &        LoveLetter &  0.21 (0.04) &  -0.58 (0.08) &   20.92 (0.56) &   731.94 (0.59) \\
         OSLA & 4 &  ExplodingKittens &  0.22 (0.04) &  -0.56 (0.08) &   21.10 (0.91) &  1226.90 (1.15) \\
          
\hline
\end{tabular}

\label{tab:benchmark_both}
\end{center}
\end{table*}


\subsection{Tic Tac Toe}
\myparagraph{Game:} Tic Tac Toe is the simplest game in TAG. The game is played on a $3 \times 3$ grid board with the players alternatively filling in the cells with their symbols (either X or O). The game ends when one of the players wins by getting $3$ matching symbols in either a horizontal, vertical or diagonal line; or when the whole board is filled, resulting in a draw.

\myparagraph{Observation and Action spaces:} In our experiments, the observation space was a $9$-dimensional vector representing the flattened board. The action space is a $9$-dimensional vector corresponding to placing the player's symbol in the corresponding cells. The action mask filters the occupied cells from the action space.

\myparagraph{Results:} Figure~\ref{fig:2p_win} (right) shows the running mean win-rates against the baseline agent. Against Random, PPO reaches a win rate over 90\%, with occasional ties. Against OSLA, PPO is able to learn a strategy that leads to winning all games.

\subsection{Diamant}

\myparagraph{Game:} ``Diamant'' is a \textit{push your luck} game for $2$ to $5$ players. Players explore caves, placing tiles with treasures or traps, and at each turn they may decide to continue exploring, or return to camp. If they return to camp, they can keep their gained treasures. If they keep exploring, they may gain further treasures; however, they may discover a trap instead, which causes every player still in the cave to lose all progress and return to the camp empty-handed. 

\myparagraph{Observation and Action spaces:} The observation space is a vector representing the tile counters, the number of gems on the current tile, and the number of gems banked in the cave. 
The action space is a $3$-dimensional flat vector. In addition to staying in the cave or returning to camp, we provide a dummy action for the case when the player is at the camp, which allows to observe how the other players are doing.

\myparagraph{Results:} The left plots in figures~\ref{fig:2p_win} and~\ref{fig:4p_win} show the win rates in $2$- and $4$-player Diamant. PPO achieves high winning rates with a very similar performance against Random and OSLA. As Diamant is a push-your-luck game, and there are only $2$ actions when the agent is in the cave, both Random and OSLA tend to return to the cave prematurely, allowing PPO to collect the higher rewards and win the game. Occasionally, PPO goes too far into the cave, gets trapped and loses the game.

\subsection{Exploding Kittens}
\myparagraph{Game:} ``Exploding Kittens'' is a card game for $2$ to $5$ players. At each turn, players may play any number of cards from their hand, but they need to finish their turns by drawing a card. If the players draw an Exploding Kitten card, they lose the game, unless they have a defensive card in their hand. The deck contains $n-1$ Exploding Kitten cards, where $n$ is the number of players. Exploding Kittens feature reactive turn-orders; some cards may be used at any time (e.g. a \textit{Nope} card may be used to cancel the effect of a card played by another player); and some cards allow the player to take cards from other players (e.g. \textit{Favor}). These features make this game a challenge for RL algorithms.

\myparagraph{Observation and Action spaces:} The observation space for Exploding Kittens contain the card type counters for each card in the player's hand, the number of cards the opponents have in hand, the number of cards left in the draw pile and the game phase. Game phases define whether the player has a normal turn, or if it has to react to some other event, such as a \textit{Favor}, \textit{Nope} or \textit{Defuse} card. 
The action space is constructed as an action tree, but in our experiments only the flattened tree is used for action selection. This space includes ending the turn by drawing a card, playing any of the card types in hand, and all the reaction actions: picking a card type to return as a \textit{Favor}, using a \textit{Nope} card, or deciding where in the draw pile to put the exploding kitten drawn, if a \textit{Defuse} card is used.

\myparagraph{Results:} The second graphs in Figures~\ref{fig:2p_win} and~\ref{fig:4p_win} show the win-rates in Exploding Kittens. In the $2$-player setting, both versions of PPO have a good win-rate of 75\%. Interestingly, in the $4$-player setting, PPO's performance drops to 35\% and 22\% against Random and OSLA, respectively. The significant drop in performance highlights the challenge of multi-agent dynamics as the number of players increases.

\subsection{Love Letter}

\myparagraph{Game:} ``Love Letter'' is a $2$- to $4$-player card game, with the objective of gaining favour tokens by either being the last player left in the game, or having the highest numerical value card at the end of each round. It was chosen both due to its abundance of hidden information and the need of memory in tracking which player has which card.

\myparagraph{Observation and Action spaces:} For the observation space, a number of features are included: the player's current hand (hot-encoded with 1 bit for each card type), the number of each card type in the discard pile, and how many favour tokens each player has.
There are 68 actions in the action space, with 8 base actions (1 per card type). The action space grows due to the combinatorial choices some cards have, such as the \textit{Guard}, who must guess the card type of one of the opponents.

\myparagraph{Results:} The third graphs in Figures~\ref{fig:2p_win} and~\ref{fig:4p_win} show the win-rates in ``Love Letter''. In ``Love Letter'', the difference in difficulty between Random and OSLA is more evident than in the other games. Against Random, PPO has a win rate of $93\%$ and $59\%$ on $2$- and $4$-player versions, respectively. Against OSLA, the performance drops significantly, resulting in a $52\%$ win-rate with $2$ players and $21$\% with $4$ players. 



\subsection{Stratego}
\myparagraph{Game:} ``Stratego'' is a $2$-player game played on a $10 \times 10$ grid map. A player wins by either capturing their opponent's flag, or if their opponent is unable to perform any actions. The game features hidden information: the type of each unit is initially known only by the player that owns it, only becoming public information if it is involved in combat. 

\myparagraph{Observation and Action spaces:} The observation space is an encoding of the $10 \times 10$ grid board, with hot-encoded unit types. $27$ feature maps each represent a separate player-exclusive unit type. 
The action space for ``Stratego'' is substantially larger than the other games implemented, reaching a maximum of $4400$ actions available for the agent. This large action space is due to each player having $40$ units, as well as the scout unit's ability to travel to any non-diagonal unoccupied grid on the board.

\myparagraph{Results:} Figure~\ref{fig:stratego} (left) presents the training results on ``Stratego''. The ``Stratego'' implementation in TAG declares a draw if a game is unfinished within $800$ steps ($400$ per agent). Without this restriction some episodes may last over $2000$ steps without reaching the end. 
The plots of the games' lengths (Figure~\ref{fig:stratego}, center) and rewards (right) show that most games ended in a tie, with episode lengths around $400$ steps, while the rewards remain close to $0$. 
Against Random, PPO shows quick improvements over time, with clear indications of learning a good policy. In contrast, OSLA is skilled at avoiding a loss, making games often end in a tie, without PPO showing high rewards throughout training. ``Stratego'' is challenging to learn due to the long episode lengths and sparse rewards, and it's possible that longer training times would improve learning. 

\section{Challenges and Opportunities}


This section describes the challenges that Modern TTGs present to RL agents, and the opportunities PyTAG brings.

\subsection{Observation Space}

\subsubsection{Representation}

Most RL benchmarks focus on having simple structured observations, either working with a set of scalar values, or using images. TTGs, however, use various components (e.g. cards, dice, boards) to provide information to players. The number and properties of these components vary widely from game to game, and thus make representing the observation space much more challenging. For example:

\begin{itemize}
    \item \textbf{Cards} can represent resources (``Settlers of Catan''), or actions the player can perform (``Exploding Kittens''). Some games, such as ``Dominion'' and ``Terraforming Mars'', have large amounts of unique cards, which may feature complex descriptions of rules in natural language. 
    \item \textbf{Boards} may take arbitrary shapes without any restrictions, including regular grids (chess, go), graphs (``Pandemic''), or hexagonal grids (``Settlers of Catan''). While some boards are fixed, others get built from reusable tiles (``Descent'', ``Gloomhaven'', ``Carcassonne'').
\end{itemize}

In the Java interface of TAG, these are represented as objects, accessible to the AI agents directly. However, RL agents require fixed and specially-tailored observation spaces, easily translated from Java to Python. PyTAG brings the creation of game-specific wrappers for compatibility with RL agents on multiple games of varying complexity, which preserves all spatial and relational information in the data.
We therefore open up the opportunity of research into efficiently exploring complex game spaces, via the observations from the large collection of existing games in TAG, now exposed to Python.


\subsubsection{Partial Observability}

Importantly, many TTGs include hidden information to either some players only (e.g. the player's own hand of cards is only visible to themselves) or to all players (e.g. a face-down deck of cards). Partial observability may apply to a whole component or set of components, but also to only a part of a component (e.g. in ``Stratego'', the player observes the opponent's pieces positions, but not their value). This raises the challenge for AI players to learn complete models of the world, in order to make accurate decisions with regards to the missing information.

We tested LSTMs as an approach for tackling this challenge. However, our results do not indicate an improvement in performance for the RL agents in any of the games used as testbeds. This could be due to more efficient training needed to allow the agent to learn how to use the temporal information, or the observations may contain enough information to play well enough against the baseline agents, without extra enhancements required.

Generally, training against SFP agents adds an important overhead, especially if having multiple instances of SFP agents in the same training setup. The multi-processed environments are efficient to speed up the running time of experiments, but the action selection for the SFP agent is not parallelised. We propose the opportunity of using a pool of SFP opponents in the training, which get progressively more challenging. This is hypothesised to lead to more efficient RL training, especially in the beginning, as the RL agent would get the chance to observe winning game states, as opposed to always losing against strong opponents. This can be seen as a form of curriculum learning~\cite{narvekar2020curriculum}.

We further open up the opportunity to explore more in-depth the effect of adding memory to RL agents in games with complex structures of hidden information, as well as more complex training setups against proficient opponents. While the aim of this paper is to show the feasibility of a single RL approach across a set of games with very different action and observation spaces, PyTAG offers complex environments for the study of other RL methodologies, such as 'self-play'~\cite{hernandez2019generalized}.

\subsubsection{Stochasticity}

TTGs provide a large source of stochasticity in various forms. Games typically have randomly-shuffled decks of cards, dice, random board generative processes, or player choices during setup, usually for replayability purposes. This results in each game played starting from a unique point in the game state space. Such actions are often repeated throughout the game, leading to very different gameplay experiences. Consequently, RL agents using TTGplay data require high efficiency for training, due to the very low probability of observing the same game state repeatedly.
PyTAG can serve as a benchmark for generalisation, as AI agents try to learn handling efficiently large, varied and dynamic state spaces.


\subsection{Action Spaces}

\subsubsection{Size and Structure}

Reinforcement Learning agents typically need to know the number of all possible actions in a game, in any game state, before learning begins, with these often architecturally encoded as one output neuron per possible action. Most TTGs, however, feature dynamic action spaces that are highly dependent on the current game state. For example, in ``Settlers of Catan'', the actions available when trading with other players are from a completely different subsection of the action space than those available when deciding what structures to build on the map. Further, the number of possible actions in some states may reach millions in TAG, due to the combinatorial possibilities of some actions and the very large and complex action and state spaces. 
TAG dynamically computes which actions are available in each state. While this is sufficient for SFP agents, it is unsuitable for RL agents, due to their requirement for a fixed action space.



In PyTAG, we adopt a similar methodology to \cite{bamford2021generalising}, by building structured action trees of the complete action space, allowing for full specification of action types and properties. Further, we apply action masking in order to filter out illegal actions depending on the given game state. 
TTGs in TAG do feature highly dynamic action spaces, with many different types of actions, all with different properties, much different to other environments readily available for RL: for example, video games are restricted to the mapping of keys on an input controller to in-game actions, which TTGS do not abide to. We therefore suggest a research direction for handling complex action spaces efficiently, most similar to the dynamic and varied nature of real-life interactions and problems.



\subsection{Rewards}

Another challenge comes from choosing the right reward function to use. 
Recent work in TAG has investigated the impact that changing the target objective (reward) function has in games in the TAG framework~\cite{Goodman2023}. Using the simple win/loss reward works well in shorter games, but some games may take thousands of decisions until a terminal state is reached. 
In these cases, it is generally advantageous to use a combination of the game score and game win/loss signals. 
While win condition and scoring of modern TTGs differ across games, a majority of games in the genre feature ongoing game scores, with the highest score at the game end determining the winner. This score can therefore provide a useful proxy for a player's progress towards winning (or losing) a game.

The winning condition in some games is tied to certain events, such as reaching the highest score (``Diamant''), reaching a target score (``Settlers of Catan''), destroying the opponent's units (``Stratego''), or outliving your opponents (``Exploding Kittens''). TAG allows defining game-specific heuristic functions, in addition to a game's natural scoring, to help SFP agents more easily find and explore the most promising areas of the search space. These heuristics are built with expert knowledge and include heavy biases that limit the play-styles exhibited by the AI agents as a result. Automatically defining appropriate and generally applicable reward functions, which allow the AI agent to learn how to play a game efficiently outside of the bounds of human limits and expertise, remains an open challenge.

Further, an issue that arises often is credit assignment, especially in cooperative games like ``Pandemic'': if players end up in a game state that brings them closer to winning, which of their actions should be credited with the reward signal? When facing large, complex, dynamic and stochastic TTGs, it is crucial to identify which move made was particularly successful (or detrimental) in the course of the game. As such, PyTAG allows for in-depth research into the credit assignment problem across various environments of varying complexity.



\subsection{Multi-Agent Dynamics}
TTGs typically focus on the interaction of two or more players. The games implemented in TAG range from 2 players to more than 10 (e.g. Poker, ``UNO''). Most games are competitive, but some games, such as ``Pandemic'', are collaborative and require the agents to work together in order to win the game (all players win or lose together). The presence of multiple intelligent entities acting upon the same environment adds extra noise to an agent's learning and decision-making process. PyTAG, therefore, offers the opportunity of studying RL applications in complex multi-agent domains, where they must learn how to react and adapt to other agents' play-styles, either competing or cooperating to achieve victory.

\subsection{PyTAG for Education}

PyTAG aims to open up TAG to more researchers, by offering language alternatives and wrappers for common interfaces such as OpenAI's gym. This allows new researchers to easily test existing algorithms within the large collection of TTGs in TAG, as well as easily create a variety of agents to be used for better playtesting and balancing TTGs. The number of games implemented in TAG is quickly growing, with multiple games actively being implemented, which serves as an ever increasing platform for RL research, raising multiple challenges unique to the domain of TTGs.
Last but not least, we highlight PyTAG as a more accessible pathway into the study of tabletop games. This is highly beneficial for educational environments, as the learning curve for python is lower than Java for beginner programmers. Many PhD, masters and undergraduate students are already using TAG for their university projects across several institutions, including Queen Mary University of London (UK), Malmo University (Sweden) and Leiden University (The Netherlands). PyTAG will increase the accessibility of the framework and serve as an introduction to the study of AI in complex TTGs.


\section{Conclusions}

This paper presents PyTAG: a python interface for the Tabletop Games framework (TAG). The paper demonstrates the use of the bridge between TAG and PyTAG by testing Reinforcement Learning agents in several games of varying complexity, from Tic Tac Toe to ``Exploding Kittens'' and ``Stratego''. We trained a Proximal Policy Optimisation algorithm with and without Long-Short Term Memory on several games, in $2$ and $4$ player settings. Our results showed good learning progress against the baseline agents in the framework. 

PyTAG was designed with flexibility in mind: the interface can not only be used for RL, but it also allows manipulating variables and executing functions directly on the Java side. To access a new game from TAG in Python, the user would simply need to implement two interfaces in Java for handling the observation and the action spaces, while following the given guidance on the Python APIs.

Additionally, the paper looks into some of the challenges Tabletop Games present for Reinforcement Learning research, and discusses many opportunities that this work opens up for in-depth studies of RL agents on complex TTGs, with direct benefits to the education sector, as well as to the improvement TTG play-testing and balancing processes.
 


\section{Acknowledgements}

For the purpose of open access, the author(s) has applied a Creative Commons Attribution (CC BY) license to any Accepted Manuscript version arising. This work was supported by the EPSRC IGGI CDT (EP/S022325/1). 

\bibliography{ref}
\bibliographystyle{IEEEtran}

\end{document}